\documentclass[12pt,twoside]{article}
\usepackage{algorithmic,algorithm}
\usepackage{graphicx}
\begin{document}
\newenvironment{keywords}{\centerline{\bf\small
Keywords}\begin{quote}\small}{\par\end{quote}\vskip 1ex}
\def\beq{\begin{equation}}    \def\eeq{\end{equation}}
\def\beqn{\begin{displaymath}}\def\eeqn{\end{displaymath}}
\def\bqa{\begin{eqnarray}}    \def\eqa{\end{eqnarray}}
\def\bqan{\begin{eqnarray*}}  \def\eqan{\end{eqnarray*}}

\def\l{\lambda}
\def\g{\gamma}
\def\Q{\mbox{Q}}
\def\TD{\mbox{TD}}
\def\HL{\mbox{HL}}
\def\HLQ{\mbox{HLQ}}
\def\HLS{\mbox{HLS}}
\def\sw{s}
\def\st{{s_t}}
\def\su{{s_{t+1}}}
\def\req#1{(\ref{#1})}
\def\EV{\overline{V}}
\def\E{\mathbf{E}}
\def\pri#1{{\it{Private: #1}}}


\title{\bf\Large\hrule height5pt \vskip 4mm
Temporal Difference Updating \\ without a Learning Rate
\vskip 4mm \hrule height2pt}

\author{\\
{\bf Marcus Hutter}\\[3mm]
\normalsize RSISE$\,$@$\,$ANU and SML$\,$@$\,$NICTA,
\normalsize Canberra, ACT, 0200, Australia \\
\normalsize \texttt{marcus@hutter1.net \ \ \ \ www.hutter1.net} \\
\and \\
{\bf Shane Legg}\\[3mm]
\normalsize IDSIA, Galleria 2, Manno-Lugano CH-6928, Switzerland\\
\normalsize \texttt{shane@vetta.org \ \ \ \ www.vetta.org/shane}
}

\date{31 October 2008}
\maketitle

\begin{abstract}
We derive an equation for temporal difference learning from
statistical principles.  Specifically, we start with the variational
principle and then bootstrap to produce an updating rule for
discounted state value estimates.  The resulting equation is similar
to the standard equation for temporal difference learning with
eligibility traces, so called $\TD(\l)$, however it lacks the
parameter $\alpha$ that specifies the learning rate.  In the place
of this free parameter there is now an equation for the learning
rate that is specific to each state transition.  We experimentally
test this new learning rule against $\TD(\l)$ and find that it
offers superior performance in various settings.  Finally, we make
some preliminary investigations into how to extend our new temporal
difference algorithm to reinforcement learning.  To do this we
combine our update equation with both Watkins' $\Q(\l)$ and
Sarsa($\l$) and find that it again offers superior performance
without a learning rate parameter.
\end{abstract}


\newpage
\section{Introduction} \label{sec:intro}

In the field of reinforcement learning, perhaps the most popular way
to estimate the future discounted reward of states is the method of
\emph{temporal difference learning}.  It is unclear who exactly introduced
this first, however the first explicit version of temporal difference
as a learning rule appears to be Witten \cite{Witten:77}.  The idea is
as follows: The \emph{expected future discounted reward} of a state
$s$ is,
\beqn
  \EV_s := \E \left \{ r_k + \g r_{k+1} + \g^2 r_{k+2} + \cdots | s_k = s \right \},
\eeqn
where the rewards $r_k, r_{k+1}, \ldots$ are geometrically discounted
into the future by $\g<1$.  From this definition it follows that,
\beq \label{eqn:trueV2}
  \EV_s = \E \left \{ r_k + \g \EV_{s_{k+1}} | s_k = s \right \}.
\eeq
Our task, at time $t$, is to compute an estimate $V^t_s$ of
$\EV_s$ for each state~$s$.  The only information we have to base
this estimate on is the current history of state transitions, $s_1,
s_2, \ldots, s_t$, and the current history of observed rewards, $r_1,
r_2, \ldots, r_t$.  Equation~\req{eqn:trueV2} suggests that at time
$t+1$ the value of $r_t + \g V_{s_{t+1}}$ provides us with information
on what $V^t_s$ should be: If it is higher than $V^t_{s_t}$ then
perhaps this estimate should be increased, and vice versa.  This
intuition gives us the following estimation heuristic
for state $s_t$,
\beqn
  V^{t+1}_{s_t} := V^t_{s_t} + \alpha \left( r_t + \g V^t_{s_{t+1}} - V^t_{s_t} \right),
\eeqn
where $\alpha$ is a parameter that controls the rate of learning.
This type of temporal difference learning is known as $\TD(0)$.

One shortcoming of this method is that at each time step the value of
only the last state $s_t$ is updated.  States before the last state
are also affected by changes in the last state's value and thus these
could be updated too.  This is what happens with so
called \emph{temporal difference learning with eligibility traces},
where a history, or trace, is kept of which states have been recently
visited.  Under this method, when we update the value of a state we
also go back through the trace updating the earlier states as well.
Formally, for any state $s$ its eligibility trace is computed by,
\beqn
  E^t_s := \left\{ \begin{array}{ll}
  \g \l E^{t-1}_s & \textrm{ if $s \neq s_t$,}\\
  \g \l E^{t-1}_s + 1 & \textrm{ if $s = s_t$,}
\end{array} \right.
\eeqn
where $\l$ is used to control the rate at which the eligibility trace
is discounted.  The temporal difference update is then, for all states
$s$,
\beq\label{eq:tdl}
  V^{t+1}_s := V^t_s + \alpha E^t_s \left( r + \g V^t_{s_{t+1}} - V^t_{s_t} \right).
\eeq
This more powerful version of temporal different learning is known as
$\TD(\l)$ \cite{Sutton:88}.

The main idea of this paper is to derive a temporal difference rule
from statistical principles and compare it to the standard heuristic
described above.
Superficially, our work has some similarities to LSTD($\lambda$)
(\cite{Lagoudakis:03} and references therein).  However LSTD is
concerned with finding a least-squares linear function approximation,
it has not yet been developed for general $\gamma$ {\em and}
$\lambda$, and has update time quadratic in the number of
features/states. On the other hand, our algorithm ``exactly''
coincides with TD/Q/Sarsa($\lambda$) for finite state spaces, but with
a novel learning rate derived from statistical principles. We
therefore focus our comparison on TD/Q/Sarsa.  For a recent survey of
methods to set the learning rate see \cite{george:06}.

In \emph{Section~\ref{sec:derive}} we derive a least squares estimate
for the value function.  By expressing the estimate as an incremental
update rule we obtain a new form of $\TD(\l)$, which we call
$\HL(\l)$.
In \emph{Section~\ref{sec:lmp}} we compare $\HL(\l)$ to $\TD(\l)$ on a
simple Markov chain.  We then test it on a random Markov chain
in \emph{Section~\ref{sec:rmp}} and a non-stationary environment
in \emph{Section~\ref{sec:nsmp}}.  In
\emph{Section~\ref{sec:mdp}} we derive two new methods for policy learning
based on $\HL(\l)$, and compare them to Sarsa($\l$) and Watkins'
$\Q(\l)$ on a simple reinforcement learning
problem.  \emph{Section~\ref{sec:con}} ends the paper with a summary
and some thoughts on future research directions.

\section{Derivation}\label{sec:derive}

The \emph{empirical future discounted reward} of a state $s_k$ is the
sum of \emph{actual rewards} following from state $s_k$ in time steps
$k, k+1, \ldots$, where the rewards are discounted as they go into the
future.  Formally, the empirical value of state $s_k$ at time $k$ for
$k=1,...,t$ is,
\beq \label{eqn:empDef}
v_k := \sum_{u=k}^\infty \g^{u-k} r_u,
\eeq
where the future rewards $r_u$ are geometrically discounted by $\g<1$.
In practice the exact value of $v_k$ is always unknown to us as it
depends not only on rewards that have been already observed, but also
on unknown future rewards.  Note that if $s_m = s_n$ for $m \neq n$,
that is, we have visited the same state twice at different times $m$
and $n$, this does not imply that $v_n = v_m$ as the observed rewards
following the state visit may be different each time.

Our goal is that for each state $s$ the estimate $V_s^t$ should be as
close as possible to the true expected future discounted reward
$\EV_s$.  Thus, for each state $s$ we would like $V_s$ to be close
to $v_k$ for all $k$ such that $s = s_k$.  Furthermore, in
non-stationary environments we would like to discount old evidence by
some parameter $\l \in (0,1]$.  Formally, we want to minimise the
loss function,
\beq\label{eqn:loss}
L := \frac{1}{2} \sum_{k=1}^t \l^{t-k} \big( v_k -V_{s_k}^t \big)^2.
\eeq
For stationary environments we may simply set $\l = 1$ a priori.

As we wish to minimise this loss, we take the partial derivative with
respect to the value estimate of each state and set to zero,
\beqn
  \frac{\partial L}{\partial V_\sw^t}
  =  - \sum_{k=1}^t \l^{t-k} \big( v_k - V_{s_k}^t \big) \delta_{s_k\sw}\\
  =  V_\sw^t \sum_{k=1}^t \l^{t-k} \delta_{s_k\sw} - \sum_{k=1}^t \l^{t-k} \delta_{s_k\sw} v_k  = 0,
\eeqn
where we could change $V_{s_k}^t$ into $V_\sw^t$ due to the presence
of the Kronecker $\delta_{s_k\sw}$, defined $\delta_{xy} := 1$ if
$x=y$, and 0 otherwise.  By defining a discounted state visit counter
$N_s^t := \sum_{k=1}^t \l^{t-k} \delta_{s_k s}$ we get
\beq\label{eqn:Vt}
  V_\sw^t N_\sw^t = \sum_{k=1}^t \l^{t-k} \delta_{s_k\sw} v_k.
\eeq
Since $v_k$ depends on future rewards $r_k$, Equation~\req{eqn:Vt} can
not be used in its current form.
Next we note that $v_k$ has a self-consistency property with respect
to the rewards.  Specifically, the tail of the future discounted
reward sum for each state depends on the empirical value at time $t$
in the following way,
\beqn
  v_k = \sum_{u=k}^{t-1} \g^{u-k} r_u + \g^{t-k} v_t.
\eeqn
Substituting this into Equation~\req{eqn:Vt} and exchanging the order
of the double sum,
\bqan
  V_\sw^t N_\sw^t & = &
  \sum_{u=1}^{t-1} \sum_{k=1}^u \l^{t-k} \delta_{s_k\sw} \g^{u-k} r_u
  + \sum_{k=1}^t \l^{t-k} \delta_{s_k\sw} \g^{t-k} v_t \\
  & = & \sum_{u=1}^{t-1}  \l^{t-u} \sum_{k=1}^u (\l \g)^{u-k} \delta_{s_k\sw} r_u
  + \sum_{k=1}^t (\l \g)^{t-k} \delta_{s_k\sw} v_t \\
  & = & R_\sw^t + E_\sw^t v_t,
\eqan
where $E_s^t := \sum_{k=1}^t (\l \g)^{t-k} \delta_{s_k s}$ is
the eligibility trace of state $s$, and $R_s^t := \sum_{u=1}^{t-1} \l^{t-u}
E_s^u r_u$ is the discounted reward with eligibility.

$E_s^t$ and $R_s^t$ depend only on quantities known at time $t$.
The only unknown quantity is $v_t$,
which we have to replace with our current estimate of this value at
time $t$, which is $V_\st^t$.  In other words, we bootstrap our estimates.
This gives us,
\beq\label{eqn:Vt2}
  V_\sw^t N_\sw^t = R_\sw^t + E_\sw^t V_\st^t.
\eeq
For state $\sw = \st$, this simplifies to
$V_\st^t = R_\st^t /(N_\st^t - E_\st^t)$.
Substituting this back into Equation~\req{eqn:Vt2} we obtain,
\beq\label{eqn:Vt3}
  V_\sw^t N_\sw^t =  R_\sw^t + E_\sw^t \frac{R_\st^t}{N_\st^t - E_\st^t}.
\eeq
This gives us an explicit expression for our $V$ estimates.  However,
from an algorithmic perspective an incremental update rule is more
convenient.  To derive this we make use of the relations,
\beqn
  N_s^{t+1} = \l N_s^t + \delta_{\su s},\qquad
  E_s^{t+1} = \l \g E_s^t + \delta_{\su s},\qquad
  R_s^{t+1} = \l R_s^t + \l E_s^t r_t,
\eeqn
with $N_s^0=E_s^0=R_s^0=0$.
Inserting these into Equation~\req{eqn:Vt3} with $t$ replaced by $t+1$,
\bqan
  V_\sw^{t+1} N_\sw^{t+1}
  & = & R_\sw^{t+1} + E_\sw^{t+1} \frac{R_\su^{t+1}} {N_\su^{t+1} - E_\su^{t+1}}\\
  & = & \l R_\sw^t + \l E_\sw^t r_t +
  E_\sw^{t+1} \frac{R_\su^t \! + E_\su^t r_t}{ N_\su^t \! - \g E_\su^t}.
\eqan

By solving Equation~\req{eqn:Vt2} for $R_s^t$ and substituting
back in,
\beqn
  V_\sw^{t+1} N_\sw^{t+1} = \l \big( V_\sw^t N_\sw^t - E_\sw^t V_\st^t \big) + \l E_\sw^t r_t + E_\sw^{t+1}
  \frac{N_\su^t V_\su^t \! - E_\su^t V_\st^t \! + E_\su^t r_t}{N_\su^t - \g E_\su^t}
\eeqn
\beqn
  = \: \big( \l N_\sw^t + \delta_{\su\sw} \big) V_\sw^t
  \! - \delta_{\su\sw} V_\sw^t \! - \l E_\sw^t V_\st^t +  \l E_\sw^t r_t
  \phantom{xxxxxx}
\eeqn
\beqn
  \phantom{xxxxxxxxx} + E_\sw^{t+1}
  \frac{N_\su^t V_\su^t \! -E_\su^t V_\st^t \! + E_\su^t r_t}{N_\su^t - \g E_\su^t}.
\eeqn
Dividing through by $N_\sw^{t+1} (=\l N_\sw^t \! +
\delta_{\su\sw})$,
\beqn
  V_\sw^{t+1} =
  V_\sw^t
  + \frac{-\delta_{\su\sw} V_\sw^t \! - \l E_\sw^t V_\st^t +  \l E_\sw^t r_t}
  {\l N_\sw^t \! + \delta_{\su\sw}} \phantom{xxxxxxxxxxxxxxx}
\eeqn
\beqn
  \phantom{xxxxxxxxx} + \frac{(\l \g E_\sw^t + \delta_{\su\sw})
  (N_\su^t V_\su^t \! -E_\su^t V_\st^t \! + E_\su^t r_t)}
  {(N_\su^t  - \g E_\su^t) (\l N_\sw^t \! + \delta_{\su\sw}) }.
\eeqn

Making the first denominator the same as the second, then expanding
the numerator,
\beqn
  V_\sw^{t+1} =
  V_\sw^t + \frac{
  \l E^t_s r_t N^t_\su - \l E^t_s V^t_\st N^t_\su - \delta_{\su s} V^t_s N^t_\su
  - \l \g E^t_\su E^t_{s} r_t}
  {(N_\su^t - \g E_\su^t) (\l N_\sw^t \! + \delta_{\su\sw}) } \phantom{xxxxxxxx}
\eeqn
\beqn
  + \frac{
    \l \g E^t_\su E^t_s V^t_\st + \g E^t_\su V^t_s \delta_{\su s}
  + \l \g E^t_s N^t_\su V^t_\su - \l \g E^t_s E^t_\su V^t_\st}
  {(N_\su^t - \g E_\su^t) (\l N_\sw^t \! + \delta_{\su\sw}) }
\eeqn
\beqn
  + \frac{\l \g E^t_s E^t_\su r_t + \delta_{\su s} N^t_\su V^t_\su
  - \delta_{\su s} E^t_\su V^t_\st + \delta_{\su s} E^t_\su r_t}
  {(N_\su^t - \g E_\su^t) (\l N_\sw^t \! + \delta_{\su\sw}) }.
\eeqn
After cancelling equal terms (keeping in mind that in every term with
a Kronecker $\delta_{xy}$ factor we may assume that $x=y$ as the term
is always zero otherwise), and factoring out $E^t_s$ we obtain,
\beqn
  V_\sw^{t+1} = V_\sw^t +
  \frac{ E^t_s \big(
  \l r_t N^t_\su - \l V^t_\st N^t_\su + \g V^t_s \delta_{\su s}
  + \l \g N^t_\su V^t_\su - \delta_{\su s} V^t_\st + \delta_{\su s} r_t \big)}
  {(N_\su^t - \g E_\su^t) (\l N_\sw^t \! + \delta_{\su\sw}) }
\eeqn
Finally, by factoring out $\l N_\su^t + \delta_{\su\sw}$ we obtain our
update rule,
\beq\label{eqn:new}
  V_\sw^{t+1} = V_\sw^t \, + \, E_\sw^t \: \beta_t(\sw,\su) \, \big( r_t + \g V_\su^t - V_\st^t \big),
\eeq
where the learning rate is given by,
\beq\label{eqn:newrate}
  \beta_t(\sw,\su) \: := \:
  \frac{1}{N_\su^t \! - \g E_\su^t}
  \: \: \frac{N_\su^t}{N_\sw^t}.
\eeq
Examining Equation~\req{eqn:new}, we find the usual update equation
for temporal difference learning with eligibility traces (see
Equation~\req{eq:tdl}), however the learning rate $\alpha$ has now
been replaced by $\beta_t(s,\su)$.  This learning rate was derived
from statistical principles by minimising the squared loss between the
estimated and true state value.  In the derivation we have exploited
the fact that the latter must be self-consistent and then bootstrapped
to get Equation~\req{eqn:Vt2}.  This gives us an equation for the
learning rate for each state transition at time $t$, as opposed to the
standard temporal difference learning where the learning rate $\alpha$
is either a fixed free parameter for all transitions, or is decreased
over time by some monotonically decreasing function.  In either case,
the learning rate is not automatic and must be experimentally tuned
for good performance.  The above derivation appears to theoretically
solve this problem.

The first term in $\beta_t$ seems to provide some type of
normalisation to the learning rate, though the intuition behind this
is not clear to us.  The meaning of second term however can be
understood as follows: $N^t_s$ measures how often we have visited
state $s$ in the recent past.  Therefore, if $N^t_s \ll N^t_\su$ then
state $s$ has a value estimate based on relatively few samples, while
state $\su$ has a value estimate based on relatively many samples.  In
such a situation, the second term in $\beta_t$ boosts the learning
rate so that $V_s^{t+1}$ moves more aggressively towards the
presumably more accurate $r_t + \g V_\su^t$.  In the opposite
situation when $\su$ is a less visited state, we see that the reverse
occurs and the learning rate is reduced in order to maintain the
existing value of $V_s$.

\section{A simple Markov process} \label{sec:lmp}

For our first test we consider a simple Markov process with 51 states.
In each step the state number is either incremented or decremented by
one with equal probability, unless the system is in state 0 or 50 in
which case it always transitions to state 25 in the following step.
When the state transitions from 0 to 25 a reward of 1.0 is generated,
and for a transition from 50 to 25 a reward of -1.0 is generated.  All
other transitions have a reward of 0.  We set the discount value $\g =
0.99$ and then computed the true discounted value of each state by
running a brute force Monte Carlo simulation.

We ran our algorithm 10 times on the above Markov chain and computed
the root mean squared error in the value estimate across the states at
each time step averaged across each run.  The optimal value of $\l$
for $\HL(\l)$ was 1.0, which was to be expected given that the
environment is stationary and thus discounting old experience is not
helpful.

\begin{figure}
\begin{minipage}[t]{0.49\textwidth}
\includegraphics[width=1.07\textwidth]{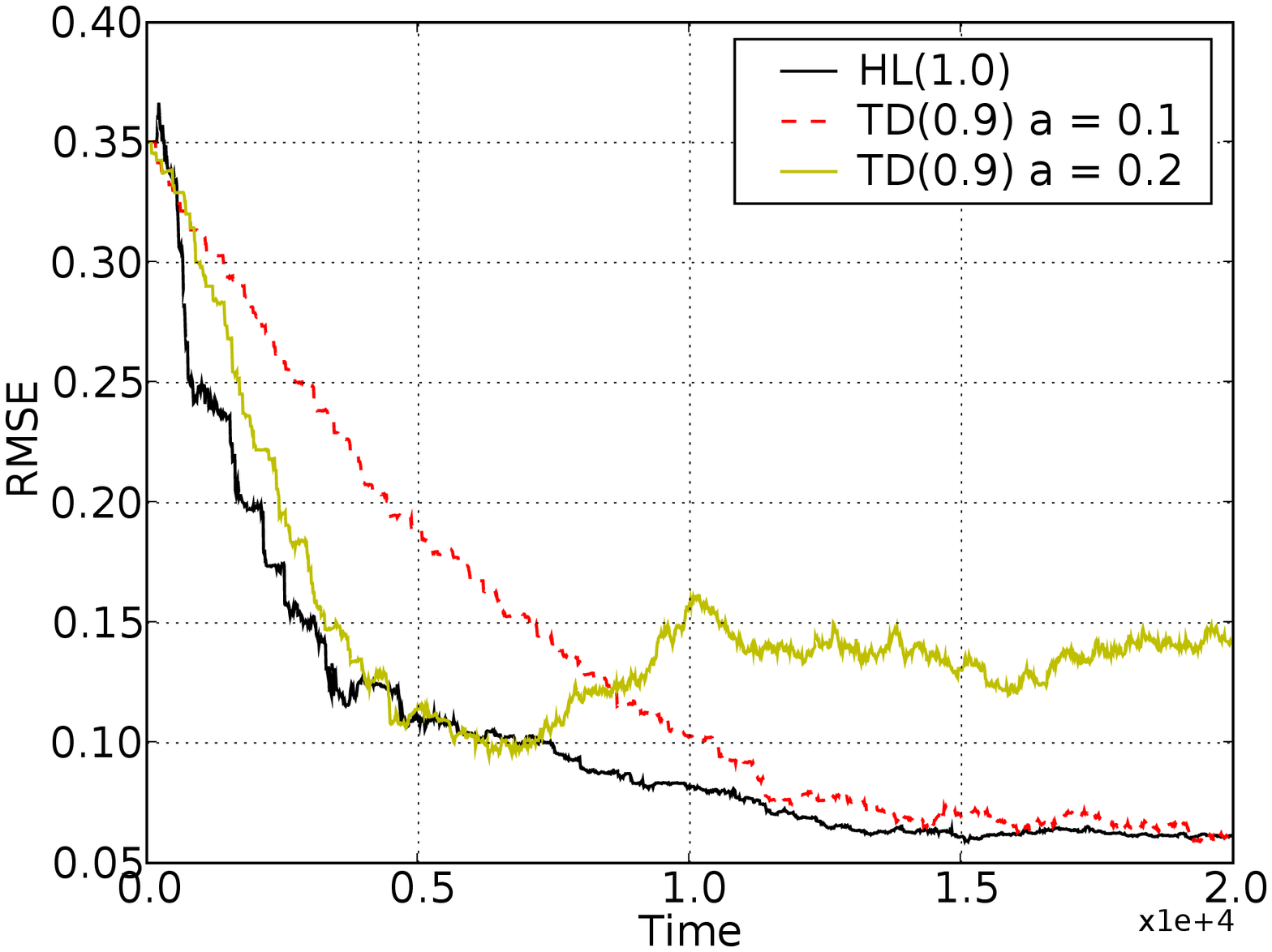}
\caption{\label{graph-1b}51 state Markov process averaged over 10 runs.
The parameter $a$ is the learning rate $\alpha$.}
\end{minipage}
\hspace{0.02\textwidth}
\begin{minipage}[t]{0.49\textwidth}
\includegraphics[width=1.07\textwidth]{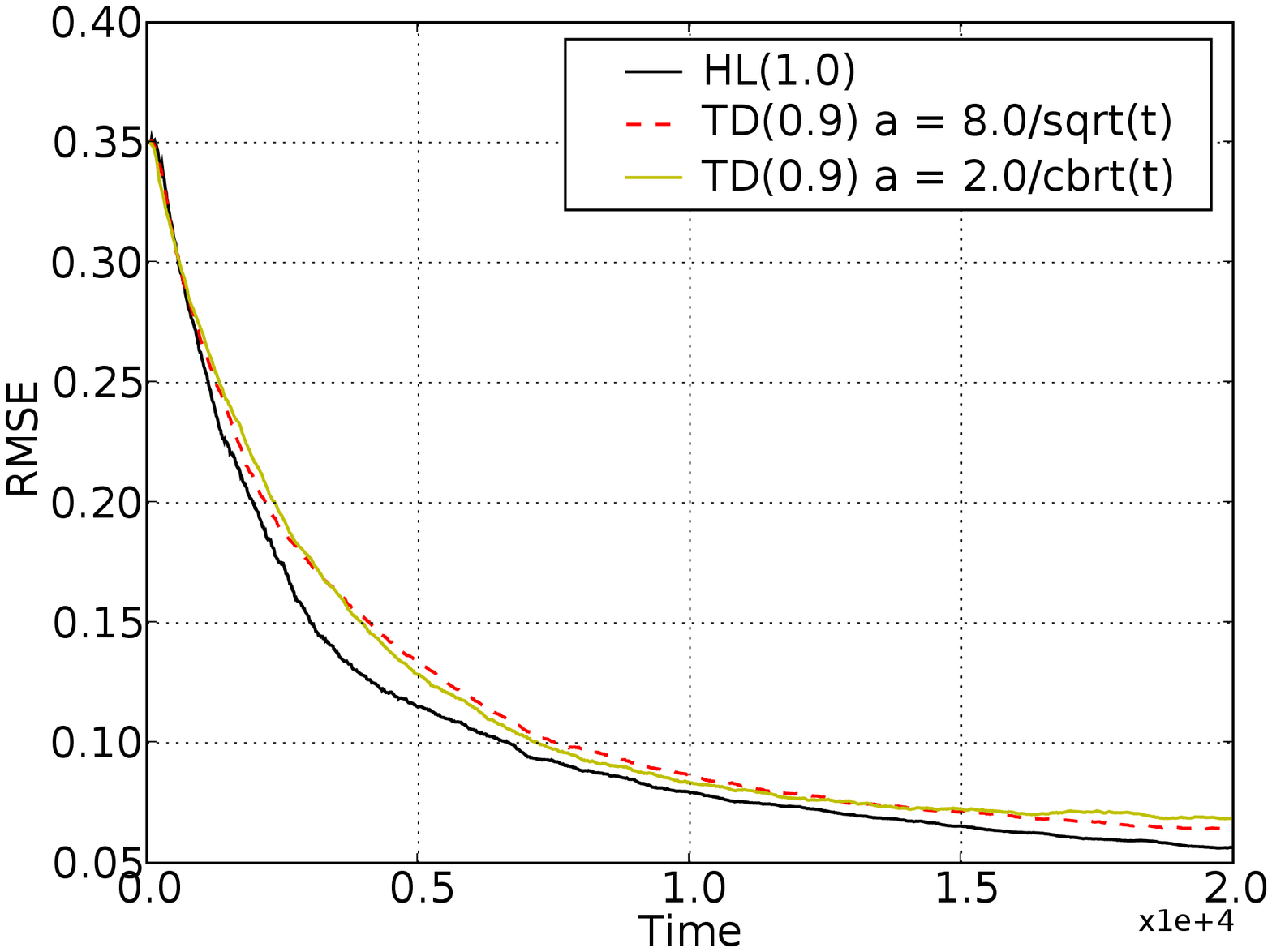}
\caption{\label{graph-1c}51 state Markov process averaged over 300 runs.}
\end{minipage}
\end{figure}

For $\TD(\l)$ we tried various different learning rates and values of
$\l$.  We could find no settings where $\TD(\l)$ was competitive with
$\HL(\l)$.  If the learning rate $\alpha$ was set too high the system
would learn as fast as $\HL(\l)$ briefly before becoming stuck.  With
a lower learning rate the final performance was improved, however the
initial performance was now much worse than $\HL(\l)$.  The results of
these tests appear in Figure~\ref{graph-1b}.

Similar tests were performed with larger and smaller Markov chains,
and with different values of~$\g$.  $\HL(\l)$ was consistently
superior to $\TD(\l)$ across these tests.  One wonders whether this
may be due to the fact that the implicit learning rate that $\HL(\l)$
uses is not fixed.  To test this we explored the performance of a
number of different learning rate functions on the 51 state Markov
chain described above.  We found that functions of the form
$\frac{\kappa}{t}$ always performed poorly, however good performance
was possible by setting $\kappa$ correctly for functions of the form
$\frac{\kappa}{\sqrt{t}}$ and $\frac{\kappa}{\sqrt[3]{t}}$.  As the
results were much closer, we averaged over 300 runs.  These results
appear in Figure~\ref{graph-1c}.

With a variable learning rate $\TD(\l)$ is performing much better,
however we were still unable to find an equation that reduced the
learning rate in such a way that $\TD(\l)$ would outperform $\HL(\l)$.
This is evidence that $\HL(\l)$ is adapting the learning rate
optimally without the need for manual equation tuning.

\section{Random Markov process} \label{sec:rmp}

To test on a Markov process with a more complex transition structure,
we created a random 50 state Markov process.  We did this by creating
a 50 by 50 transition matrix where each element was set to 0 with
probability 0.9, and a uniformly random number in the interval $[0,1]$
otherwise.  We then scaled each row to sum to 1.  Then to transition
between states we interpreted the $i^{th}$ row as a probability
distribution over which state follows state $i$.  To compute the
reward associated with each transition we created a random matrix as
above, but without normalising.  We set $\g = 0.9$ and then ran a
brute force Monte Carlo simulation to compute the true discounted
value of each state.

\begin{figure}
\begin{minipage}[t]{0.49\textwidth}
\includegraphics[width=1.07\textwidth]{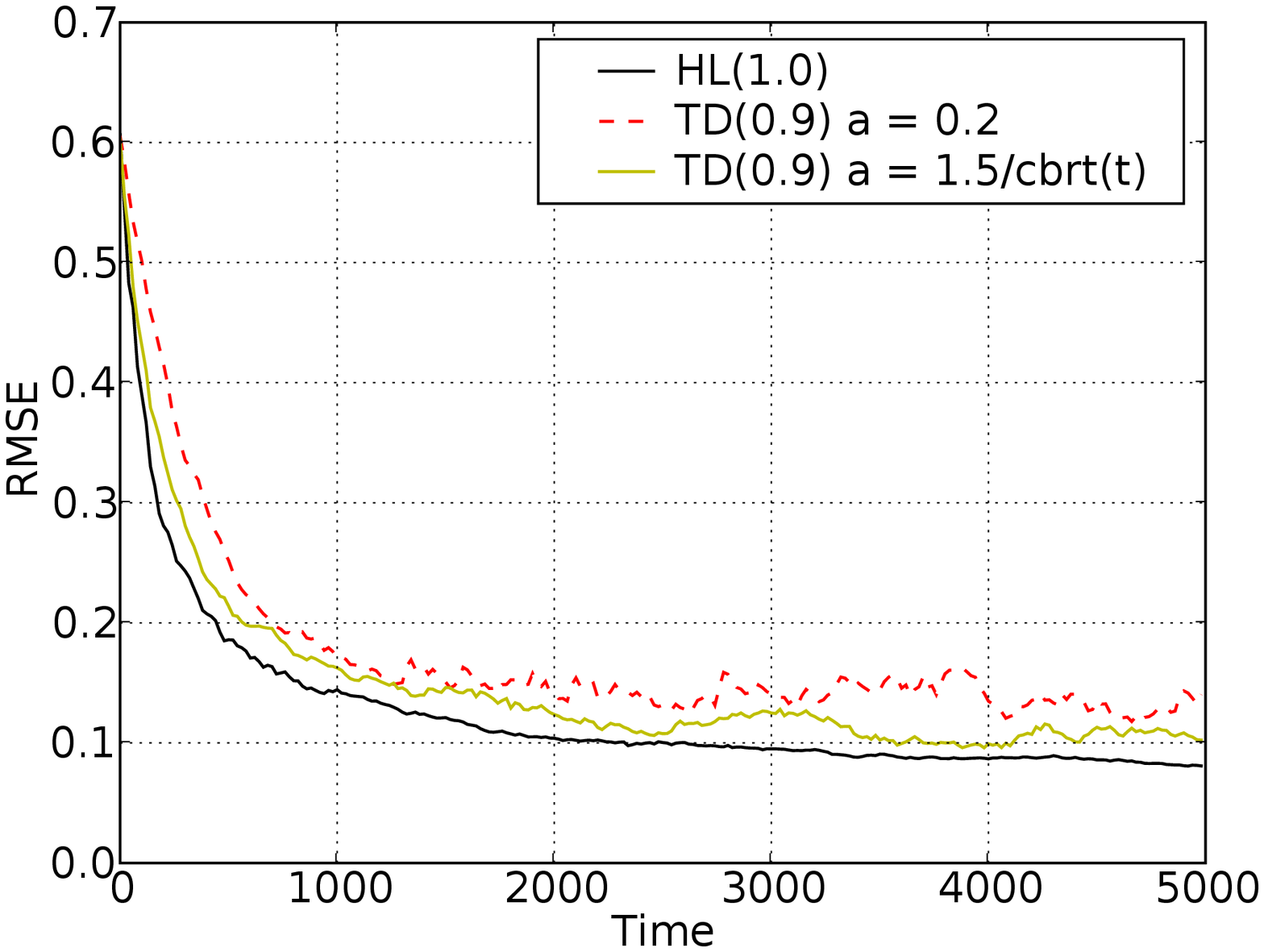}
\caption{\label{graph-1d}Random 50 state Markov process.
The parameter $a$ is the learning rate $\alpha$.}
\end{minipage}
\hspace{0.02\textwidth}
\begin{minipage}[t]{0.49\textwidth}
\includegraphics[width=1.07\textwidth]{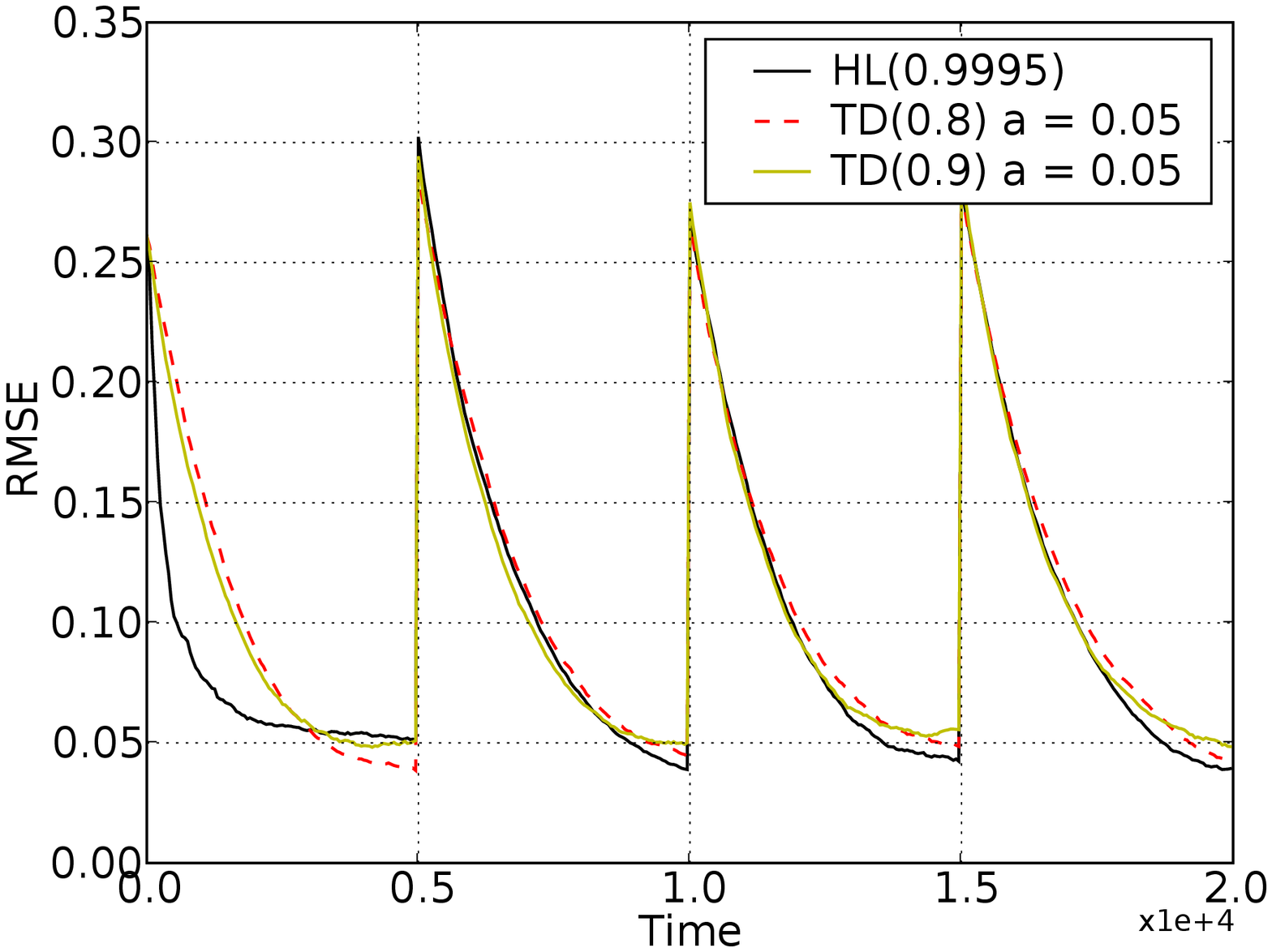}
\caption{\label{graph-2}21 state non-stationary Markov process.}
\end{minipage}
\end{figure}

The $\l$ parameter for $\HL(\l)$ was simply set to 1.0 as the
environment is stationary.  For $\TD$ we experimented with a range of
parameter settings and learning rate decrease functions.  We found
that a fixed learning rate of $\alpha = 0.2$, and a decreasing rate of
$\frac{1.5}{\sqrt[3]{t}}$ performed reasonable well, but never as well
as $\HL(\l)$.  The results were generated by averaging over 10 runs,
and are shown in Figure~\ref{graph-1d}.

Although the structure of this Markov process is quite different to
that used in the previous experiment, the results are again similar:
$\HL(\l)$ preforms as well or better than $\TD(\l)$ from the beginning
to the end of the run.  Furthermore, stability in the error towards
the end of the run is better with $\HL(\l)$ and no manual learning
tuning was required for these performance gains.

\section{Non-stationary Markov process} \label{sec:nsmp}

The $\l$ parameter in $\HL(\l)$, introduced in
Equation~\req{eqn:loss}, reduces the importance of old observations
when computing the state value estimates.  When the environment is
stationary this is not useful and so we can set $\l = 1.0$, however in
a non-stationary environment we need to reduce this value so that the
state values adapt properly to changes in the environment.  The more
rapidly the environment is changing, the lower we need to make $\l$ in
order to more rapidly forget old observations.

To test $\HL(\l)$ in such a setting, we used the Markov chain from
Section~\ref{sec:lmp}, but reduced its size to 21 states to speed up
convergence.  We used this Markov chain for the first 5,000 time
steps.  At that point, we changed the reward when transitioning from
the last state to middle state to from -1.0 to be 0.5.  At time 10,000
we then switched back to the original Markov chain, and so on
alternating between the models of the environment every 5,000 steps.
At each switch, we also changed the target state values that the
algorithm was trying to estimate to match the current configuration of
the environment.  For this experiment we set $\gamma = 0.9$.

As expected, the optimal value of $\l$ for $\HL(\l)$ fell from 1 down
to about 0.9995.  This is about what we would expect given that each
phase is 5,000 steps long.  For $\TD(\l)$ the optimal value of $\l$
was around 0.8 and the optimum learning rate was around $0.05$.  As we
would expect, for both algorithms when we pushed $\l$ above its
optimal value this caused poor performance in the periods following
each switch in the environment (these bad parameter settings are not
shown in the results).  On the other hand, setting $\l$ too low
produced initially fast adaption to each environment switch, but poor
performance after that until the next environment change.  To get
accurate statistics we averaged over 200 runs.  The results of these
tests appear in Figure~\ref{graph-2}.

For some reason $\HL(0.9995)$ learns faster than $\TD(0.8)$ in the
first half of the first cycle, but only equally fast at the start of
each following cycle.  We are not sure why this is happening.  We
could improve the initial speed at which $\HL(\l)$ learnt in the last
three cycles by reducing $\l$, however that comes at a performance
cost in terms of the lowest mean squared error attained at the end of
each cycle.  In any case, in this non-stationary situation $\HL(\l)$
again performed well.

\section{Windy Gridworld} \label{sec:mdp}

Reinforcement learning algorithms such as Watkins' $\Q(\l)$
\cite{Watkins:89} and Sarsa($\l$) \cite{Rummery:94,Rummery:95} are
based on temporal difference updates.  This suggests that new
reinforcement learning algorithms based on $\HL(\l)$ should be
possible.

For our first experiment we took the standard Sarsa($\l$)
algorithm and modified it in the obvious way to use an HL temporal
difference update.  In the presentation of this algorithm we have
changed notation slightly to make things more consistent with that
typical in reinforcement learning.  Specifically, we have dropped the
$t$ super script as this is implicit in the algorithm specification,
and have defined $Q(s,a) := V_{(s,a)}$, $E(s,a) := E_{(s,a)}$ and
$N(s,a) := N_{(s,a)}$.  Our new reinforcement learning algorithm,
which we call HLS($\l$) is given in Algorithm~\ref{alg:hls}.
Essentially the only changes to the standard Sarsa($\l$) algorithm
have been to add code to compute the visit counter $N(s,a)$, add a
loop to compute the $\beta$ values, and replace $\alpha$ with $\beta$
in the temporal difference update.

\begin{figure}
\begin{minipage}[b]{0.49\textwidth}
\includegraphics[width=\textwidth,height=15em]{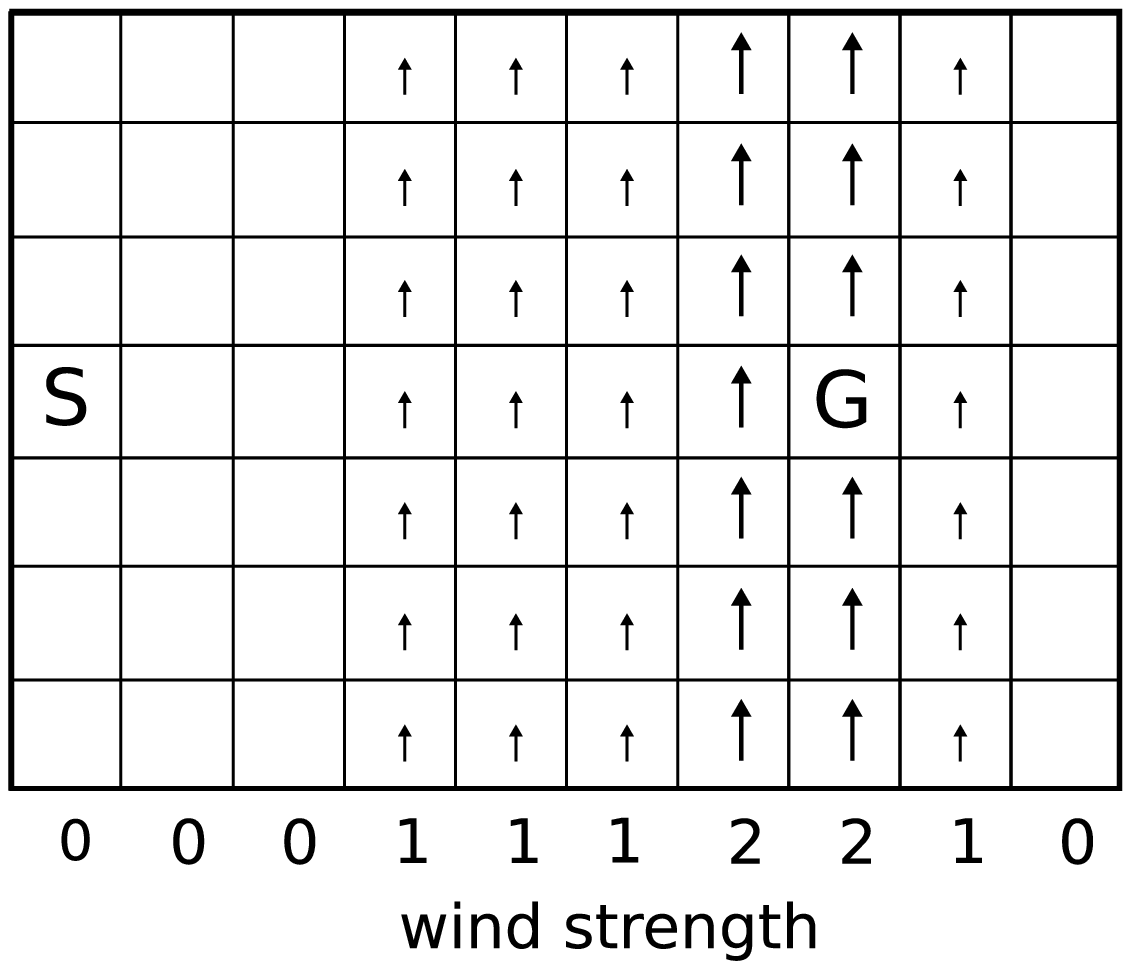}
\caption{\label{windworld}[Windy Gridworld]
S marks the start state and G the goal state, at which
the agent jumps back to S with a reward of 1.}
\end{minipage}
\hspace{0.02\textwidth}
\begin{minipage}[b]{0.49\textwidth}
\includegraphics[width=1.07\textwidth]{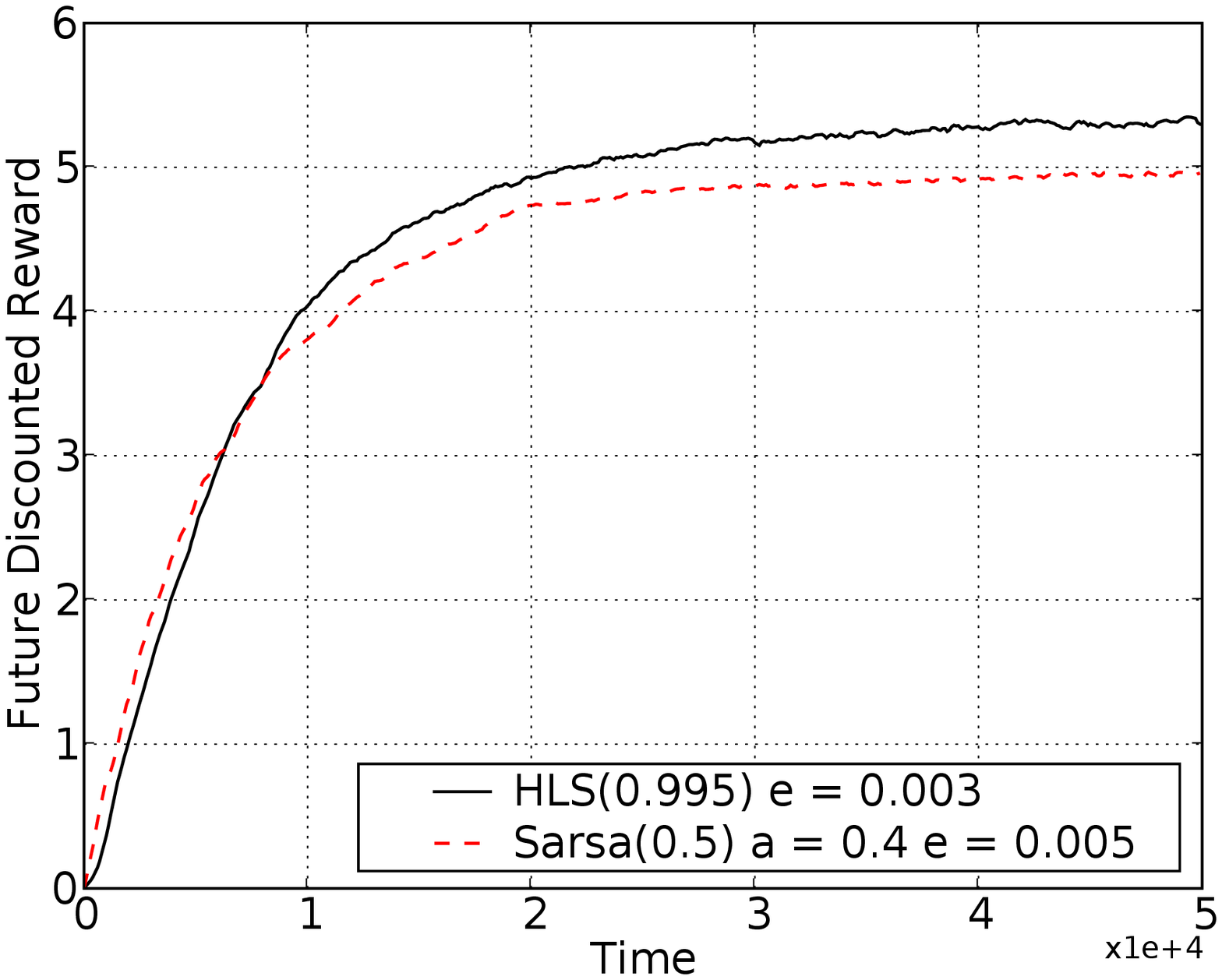}
\caption{\label{graph-3}Sarsa($\l$) vs.\ HLS($\l$) in the Windy Gridworld.
Performance averaged over 500 runs.  On the graph, $e$ represents
the exploration parameter $\epsilon$, and $a$ the learning rate
$\alpha$.}
\end{minipage}
\end{figure}

To test HLS($\l$) against standard Sarsa($\l$) we used the Windy
Gridworld environment described on page 146 of \cite{Sutton:98}.  This
world is a grid of 7 by 10 squares that the agent can move through by
going either up, down, left of right.  If the agent attempts to move
off the grid it simply stays where it is.  The agent starts in the
$4^{th}$ row of the $1^{st}$ column and receives a reward of 1 when it
finds its way to the $4^{th}$ row of the $8^{th}$ column.  To make
things more difficult, there is a ``wind'' blowing the agent up 1 row
in columns 4, 5, 6, and 9, and a strong wind of 2 in columns 7 and 8.
This is illustrated in Figure~\ref{windworld}.  Unlike in the original
version, we have set up this problem to be a continuing discounted
task with an automatic transition from the goal state back to the
start state.

\begin{algorithm}[t]
\caption{HLS($\l$)}
\label{alg:hls}
\begin{algorithmic}
\STATE Initialise $Q(s,a)=0$, $N(s,a)=1$ and $E(s,a)=0$ for all $s$, $a$
\STATE Initialise $s$ and $a$
\REPEAT
  \STATE Take action $a$, observed $r$, $s'$
  \STATE Choose $a'$ by using $\epsilon$-greedy selection on $Q(s',\cdot)$
  \STATE $\Delta \leftarrow r + \g Q(s',a') - Q(s,a)$
  \STATE $E(s,a) \leftarrow E(s,a)+1$
  \STATE $N(s,a) \leftarrow N(s,a)+1$

  \FORALL {$s,a$}
  \STATE $\beta( (s,a), (s',a') ) \leftarrow \frac{1}{N(s',a') - \g E(s',a')} \: \: \frac{N(s',a')}{N(s,a)}$
  \ENDFOR

  \FORALL {$s,a$}
    \STATE $Q(s,a) \leftarrow Q(s,a) + \beta \big( (s,a), (s',a') \big) E(s,a) \Delta$
    \STATE $E(s,a) \leftarrow \g \l E(s,a)$
    \STATE $N(s,a) \leftarrow \l N(s,a)$
  \ENDFOR
  \STATE $s \leftarrow s' ; a \leftarrow a'$
\UNTIL end of run
\end{algorithmic}
\end{algorithm}

We set $\gamma = 0.99$ and in each run computed the empirical future
discounted reward at each point in time.  As this value oscillated we
also ran a moving average through these values with a window length of
50.  Each run lasted for 50,000 time steps as this allowed us to see
at what level each learning algorithm topped out.  These results
appear in Figure~\ref{graph-3} and were averaged over 500 runs to get
accurate statistics.

Despite putting considerable effort into tuning the parameters of
Sarsa($\l$), we were unable to achieve a final future discounted
reward above 5.0.  The settings shown on the graph represent the best
final value we could achieve.  In comparison HLS($\l$) easily beat
this result at the end of the run, while being slightly slower than
Sarsa($\l$) at the start.  By setting $\l = 0.99$ we were able to
achieve the same performance as Sarsa($\l$) at the start of the run,
however the performance at the end of the run was then only slightly
better than Sarsa($\l$).  This combination of superior performance and
fewer parameters to tune suggest that the benefits of $\HL(\l)$ carry
over into the reinforcement learning setting.

Another popular reinforcement learning algorithm is Watkins'
$\Q(\l)$. Similar to Sarsa($\l$) above, we simply inserted the
$\HL(\l)$ temporal difference update into the usual $\Q(\l)$
algorithm in the obvious way.  We call this new algorithm
HLQ($\l$)(not shown).  The test environment was exactly the same as
we used with Sarsa($\l$) above.

The results this time were more competitive (these results are not
shown).  Nevertheless, despite spending a considerable amount of time
fine tuning the parameters of $\Q(\l)$, we were unable to beat
$\HLQ(\l)$.  As the performance advantage was relatively modest, the
main benefit of $\HLQ(\l)$ was that it achieved this level of
performance without having to tune a learning rate.

\section{Conclusions}\label{sec:con}

We have derived a new equation for setting the learning rate in
temporal difference learning with eligibility traces.  The equation
replaces the free learning rate parameter $\alpha$, which is normally
experimentally tuned by hand.  In every setting tested, be it
stationary Markov chains, non-stationary Markov chains or
reinforcement learning, our new method produced superior results.

To further our theoretical understanding, the next step would be to
try to prove that the method converges to correct estimates.  This can
be done for $\TD(\lambda)$ under certain assumptions on how the
learning rate decreases over time.  Hopefully, something similar can
be proven for our new method.  In terms of experimental results, it
would be interesting to try different types of reinforcement learning
problems and to more clearly identify where the ability to set the
learning rate differently for different state transition pairs helps
performance.  It would also be good to generalise the result to
episodic tasks.  Finally, just as we have successfully merged
$\HL(\l)$ with Sarsa($\l$) and Watkins' $\Q(\l)$, we would also like
to see if the same can be done with Peng's $\Q(\l)$ \cite{Peng:96}, and
perhaps other reinforcement learning algorithms.

\paragraph{Acknowledgements.}
This research was funded by the Swiss NSF grant 200020-107616.

\bibliographystyle{alpha}
\begin{small}

\end{small}
\end{document}